\documentclass{article}  
\usepackage{graphicx}  
\usepackage{hyperref}  
\usepackage{caption} 
\usepackage{float}
\usepackage{amsmath}
\usepackage{caption}

\title{Visualizing Celebrity Dynamics in Video Content: A Proposed Approach Using Face Recognition Timestamp Data}

\author{%
  \begin{minipage}{\textwidth}
    \centering
    {\large Doğanay Demir and İlknur Durgar Elkahlout}\\[4pt]
    {\normalsize
    Digital Data Analytics, Department of New Media, TRT, İstanbul, Türkiye}\\[6pt]
    \begin{tabular}{c c}
      \texttt{doganay.demir@trt.net.tr} & \texttt{ilknur.durgar@trt.net.tr}
    \end{tabular}
  \end{minipage}
}

\date{\today}  
  
\begin{document}  
  
\maketitle  
  
\begin{abstract}  
In an era dominated by video content, understanding its structure and dynamics has become increasingly important. This paper presents a hybrid framework that combines a distributed multi-GPU inference system with an interactive visualization platform for analyzing celebrity dynamics in video episodes. The inference framework efficiently processes large volumes of video data by leveraging optimized ONNX models, heterogeneous batch inference, and high-throughput parallelism, ensuring scalable generation of timestamped appearance records. These records are then transformed into a comprehensive suite of visualizations, including appearance frequency charts, duration analyses, pie charts, co-appearance matrices, network graphs, stacked area charts, seasonal comparisons, and heatmaps. Together, these visualizations provide multi-dimensional insights into video content, revealing patterns in celebrity prominence, screen-time distribution, temporal dynamics, co-appearance relationships, and intensity across episodes and seasons. The interactive nature of the system allows users to dynamically explore data, identify key moments, and uncover evolving relationships between individuals. By bridging distributed recognition with structured, visually-driven analytics, this work enables new possibilities for entertainment analytics, content creation strategies, and audience engagement studies. 
\end{abstract}

\section{Introduction}

The rapid growth of video content in the digital era has created an ever-increasing demand for tools that can analyze, understand, and visualize visual and temporal structures. Data visualization techniques offer information through visual representations \cite{steed2025interactive}. One of the main objectives is to visually represent the patterns underlying the data. These patterns can help uncover relationships between temporal segments, highlight key events, and support decision-making processes in various domains. Consequently, effective visualization approaches play a critical role in simplifying complex video data and making it more interpretable for both experts and non-experts. Videos—whether from entertainment platforms, editing and processing applications \cite{bitter2007comparison},\cite{kindlmann2015diderot},\cite{lai2017semantic},\cite{wang2013spatially},\cite{xiao2010fast}, activity recognition and content analysis \cite{zeng2019emoco},\cite{botchen2008action},\cite{chan2019motion},\cite{fan2019vista},\cite{jang2015motionflow},\cite{lu2012timeline},\cite{meghdadi2013interactive},\cite{romero2008viz},\cite{wang2017outdoor}, educational media, or social streaming—have become primary storytelling vehicles, yet analytics has often emphasized text and audio modalities; visual dynamics such as who appears on screen when, how often, and in relation to whom remain underexplored despite their importance for narrative and social context.

Recent advances in video face datasets provide new possibilities for this domain. For example, CelebV-HQ (2022) offers a large-scale video dataset featuring rich facial attribute annotations across 15,653 identities and over 35,000 high-resolution clips—enabling detailed temporal and identity-based visual analytics \cite{celebvhq2022}. More recently, FaceVid-1K (2024) delivers a multiracial, high-quality face video dataset spanning over 200,000 clips (over 1,000 hours) with fine-grained annotations such as ethnicity, age, appearance, emotion, and actions—providing robust material for exploring co-appearance and identity dynamics \cite{facevid1k2024}. 

This paper proposes a framework for visualizing and analyzing the appearances of celebrities in video content. By leveraging face recognition technology, the system identifies celebrities in each frame of a video and stores their appearances as timestamped data. This data is then transformed into a suite of intuitive visualizations, including bar charts, pie charts, line graphs, co-appearance matrices, network graphs, stacked area charts and heatmaps. These visualizations provide insights into various aspects of a video, such as the prominence of celebrities, their screen time distribution, and the relationships between individuals based on shared appearances. 

The motivation for this system arises from the need to better understand the structure of video content. For example, in a TV show or movie episode, the frequency and duration of celebrity appearances can offer insights into storytelling dynamics, character importance, and social interactions. Such insights can be valuable for content creators, media analysts, and even fans seeking to explore their favorite shows in new ways. 

This paper is intended as a framework for analyzing celebrity dynamics in video content, bridging the gap between face recognition technology and creative data visualization. While the implementation details are not the focus, the methodology and potential applications of the system are explored in depth. By presenting this idea, we aim to inspire further exploration of video analytics and open up new possibilities for understanding media through innovative visual tools.

\section{Related Work}

Visualizing temporal and relational aspects of video data has received growing attention in recent years. Several systems focus on \textbf{timeline-based exploration of video metadata}. For example, DUET investigates alternative event layouts and interactions for timeline visualizations \cite{duet2025}, while work on adaptive 360° video timelines compares multiple styles of immersive browsing \cite{adaptive360timeline2024}. VisTellAR embeds data visualizations inside short-form videos and provides a timeline to navigate them \cite{vistellar2024}, and Data Playwright supports structured authoring of data videos aligned with narration and time \cite{dataplaywright2024}.

Furthermore, 2D timestamp visualization techniques are widely used to represent temporal patterns in data, often through line charts, scatter plots, or heatmaps, enabling clear insights into trends, periodicity, or anomalies over time \cite{chen2015peakvizor},\cite{polk2014tennivis},\cite{polk2019courttime},\cite{wang2018image}. Additionally, 3D visualization techniques leverage spatial depth to provide a richer representation of data, often used in immersive environments like virtual reality. These methods enable researchers to analyze complex datasets by incorporating depth and motion, offering enhanced insights compared to traditional 2D visualizations \cite{liao2011video}, \cite{liu2009point}, \cite{serrano2019motion}.

Beyond 2D and 3D approaches, several advanced visualization techniques have emerged to address specific challenges in video metadata analysis. \textbf{Temporal visualization techniques} such as spiral plots, Gantt charts, and animated timelines are particularly effective for identifying recurring patterns or long-term trends in timestamped data. \textbf{Multivariate visualizations} like parallel coordinates or scatterplot matrices help analyze relationships between multiple features of video metadata. Meanwhile, \textbf{geospatial visualizations} integrate spatial and temporal information, visualizing the geographic context of events or interactions. These methods expand the toolkit for analyzing complex temporal and relational aspects of video data.

In parallel, \textbf{co-occurrence and co-appearance networks} have been studied in film, media, and related domains. Penta demonstrates compound graph visualization with applications to character co-occurrence networks \cite{penta2025}. Surveys of film-industry networks emphasize actor collaborations and co-appearance structures \cite{filmsurvey2024}, while recent actor-network analyses highlight temporal and relational properties of collaboration graphs \cite{actorcollab2024}. Broader studies on digital-media co-occurrence \cite{unicooccurrence2025} and even cognitive neuroscience approaches that model characters through co-occurrence over timelines \cite{pnascortex2024} provide methodological parallels.

Finally, work on \textbf{structured video event analysis} underpins many visualization pipelines. ARC-Hunyuan-Video-7B focuses on structured video comprehension with temporally precise event representations \cite{arcvideo2025}. Action co-occurrence datasets such as CO-ACT \cite{coact2024}, video relationship graph generation models like CAGNet \cite{cagnet2024}, and hierarchical co-occurrence GCNs for action recognition \cite{ehcgcn2025} all contribute techniques for extracting the timestamped and relational signals that feed co-appearance matrices and dynamic celebrity networks.

Taken together, these strands show the convergence of timeline interaction design, co-occurrence network visualization, and structured video understanding—key ingredients for analyzing celebrity dynamics across video corpora. Despite these advancements, there is limited work on integrating these techniques into a unified framework for visualizing timestamp data and co-occurrences in video content. Our work addresses this gap by proposing a visualization framework that seamlessly integrates temporal analysis, co-appearance networks, and interactive dashboards, enabling comprehensive exploration of celebrity appearances and their interactions throughout video content.

\section{Methodology}  

This section outlines the methodology employed to design and implement the proposed framework for analyzing and visualizing celebrity dynamics in video content. The system integrates three core components: a custom-built face recognition framework, an efficient data storage and retrieval pipeline, and a visualization layer designed to deliver interactive and insightful representations of data. The modular nature of the system ensures scalability, adaptability, and accessibility across different use cases.  

\subsection{Face Recognition Framework}  

\begin{figure}[!htbp]
    \centering
    \includegraphics[width=1.0\linewidth]{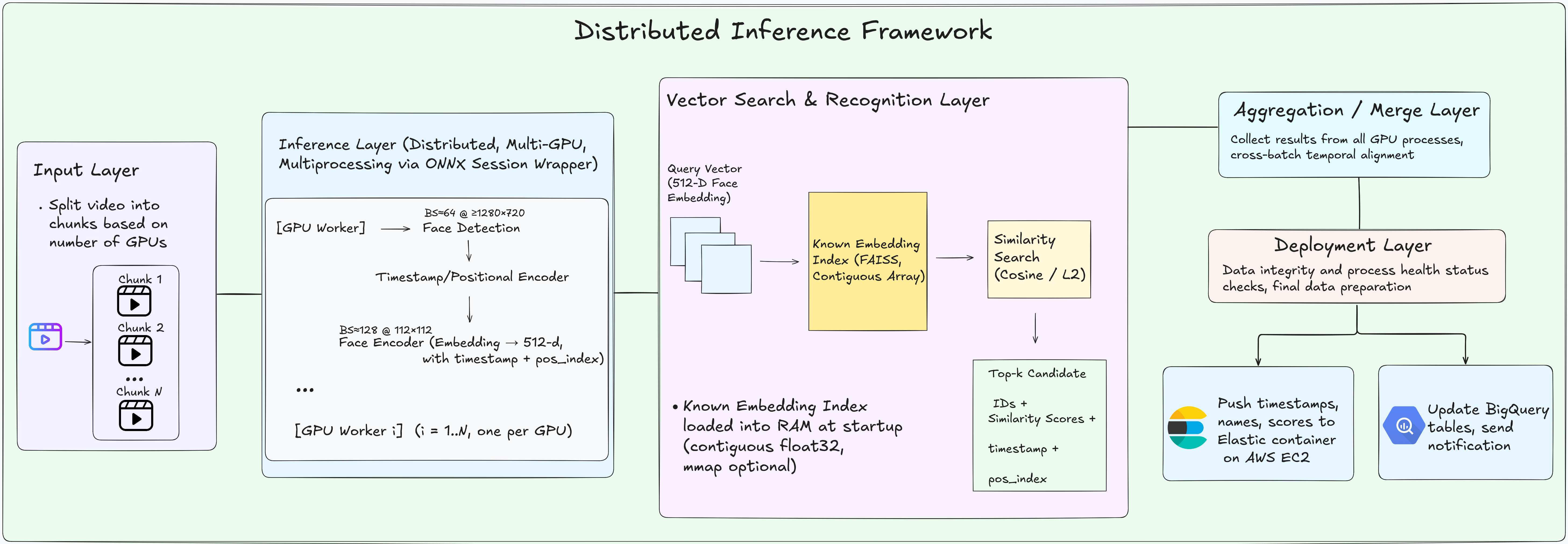}
    \captionsetup{justification=centering, font=small}
    \caption{Distributed Inference Framework Architecture}
    \label{fig:distributed_inference_framework}
    \vspace{-2mm}
\end{figure}

At the core of the system lies a proprietary \textbf{Distributed Inference Framework} developed by the authors, designed for \textbf{high-throughput celebrity recognition} in full-length TV series episodes. The framework builds on our previous work \cite{10581926} and leverages customized ONNX models, multi-GPU parallelism, and batching strategies to efficiently detect and identify individuals in video frames while preserving precise temporal information.

\subsubsection{System Architecture}

The framework is organized as a \textbf{layered architecture} with four major components:

\begin{itemize}
    \item \textbf{Input Layer:} Episodes are divided into time-based chunks, with the number of chunks matching the available GPUs. Each chunk is assigned to a GPU process, ensuring balanced parallel processing across compute resources.

    \item \textbf{Inference Layer:} Each GPU process executes the following pipeline:
    \begin{enumerate}
        \item \textbf{Face Detection:} High-resolution face detector identifies all faces in frames.
        \item \textbf{Timestamp/Positional Encoder:} Attaches timestamps and positional indices to detected faces to maintain temporal alignment across batches, addressing differing batch sizes in detection and embedding stages.
        \item \textbf{Face Encoder:} Cropped face images are transformed into 512-dimensional embeddings via ONNX models. Each embedding retains its timestamp and positional index, ensuring accurate downstream temporal analysis.
    \end{enumerate}

    This layer is executed in parallel across $N$ GPU processes, each utilizing an \textbf{ONNX Session Wrapper} for safe serialization in a multiprocessing environment. The GPU pipeline can be represented as:

    \[
    \begin{aligned}
    [\mathbf{GPU\ Worker}] &\;\rightarrow\; \text{Face Detection} \\
    &\;\rightarrow\; \text{Timestamp/Positional Encoder} \\
    &\;\rightarrow\; \text{Face Encoder (Embedding $\rightarrow$ 512-d,} \\
    &\qquad \text{with timestamp + pos\_index)}
    \end{aligned}
    \]

    For the parallel GPU workers, we denote:
    \[
    [\mathbf{GPU\ Worker}_i], \quad i = 1..N
    \]

    \item \textbf{Vector Search Layer:} Generated embeddings are compared against a precomputed \textbf{Known Embedding Index} (contiguous FAISS float32 array). Efficient similarity searches (Cosine or L2) retrieve top-$k$ closest matches, preserving timestamp information for temporal correlation.

    \item \textbf{Aggregation and Deployment:} Results from all GPU processes are merged into an \textbf{episode-level timeline}, which is pushed to AWS Elastic container storage. The BigQuery metadata table is updated to mark processed episodes, ensuring synchronized tracking across datasets.
\end{itemize}

\subsubsection{Pipeline Optimization and Automation}
The framework is optimized for scalability and high throughput:
\begin{itemize}
    \item Different batch sizes are used for detection and embedding (e.g., BS$\approx$64 and BS$\approx$128) to maximize GPU utilization.
    \item Multiprocessing ensures parallel execution of independent GPU workers.
    \item ONNX Session Wrapper allows independent, serialized model sessions per GPU process.
    \item Timestamp and positional indices maintain \textbf{temporal integrity} across batches.
\end{itemize}

\subsubsection{Integration with Visualizer and Data Systems}

\begin{figure}[!htbp]
    \centering
    \includegraphics[width=1.0\linewidth]{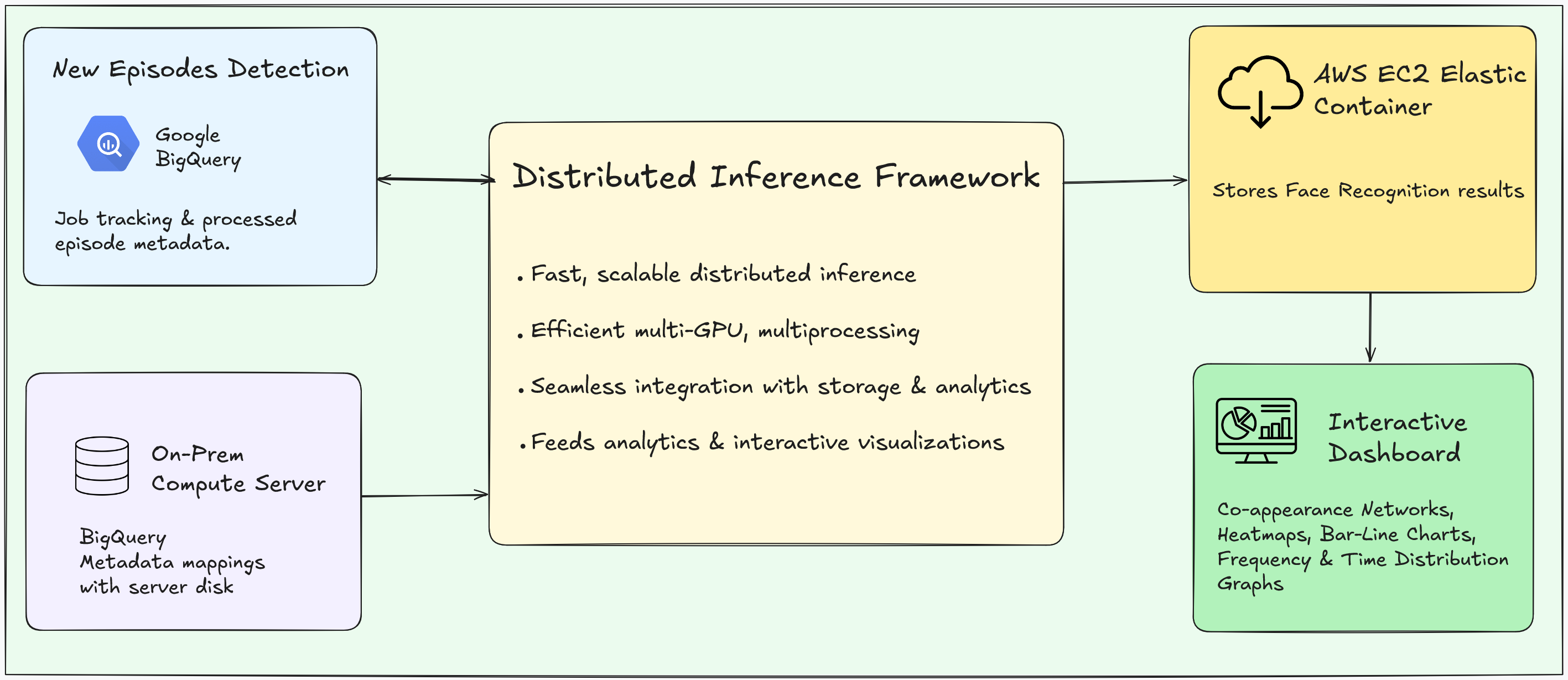}
    \captionsetup{justification=centering, font=small}
    \caption{Visualizer App Architecture}
    \label{fig:visualizer_architecture}
    \vspace{-2mm}
\end{figure}

Processed data is consumed by a \textbf{Visualizer App} that provides detailed analytics of celebrity appearances across episodes:
\begin{itemize}
    \item The \textbf{Visualizer} fetches processed data from the AWS Elastic container to generate interactive timelines, co-appearance networks, heatmaps, and seasonal comparisons.
    \item \textbf{BigQuery} stores metadata about episodes and their processing status; the Distributed Inference Framework both queries it for new episodes and updates it post-processing.
    \item \textbf{On-Prem Compute Server} maps BigQuery metadata to raw video files, where the core processing of each episode takes place.
\end{itemize}

\subsubsection{Summary}
This Distributed Inference Framework enables scalable, automated, multi-GPU celebrity recognition across full TV series episodes, preserving precise temporal information for all detected embeddings. It forms the backbone of our system, allowing downstream analytics and visualization to provide actionable insights into celebrity dynamics in video content.

\subsection{Data Storage and Retrieval}

The timestamped recognition data is stored in an Elasticsearch container hosted on an Amazon EC2 instance. Elasticsearch was chosen not only for its scalability and high-speed indexing, but also for its ability to serve as the analytical backbone for visualization. By enabling structured queries and aggregations, it directly supports the generation of charts, graphs, and co-appearance networks described in the visualization layer.

Data is organized with flexible schemas that facilitate both fine-grained lookups and higher-level summaries, including:
\begin{itemize}
\item Retrieving all appearances of a specific individual across one or multiple videos to support bar and pie chart distributions.
\item Identifying co-appearances within a defined temporal window, enabling the construction of network graphs and co-appearance matrices.
\item Generating aggregate statistics such as total screen time, frequency of appearances, and interaction patterns, which power trend analyses in line and stacked area charts.
\item Supporting temporal and segment-based queries that underpin heatmaps and seasonal comparisons for longitudinal insights.
\end{itemize}

Access is secured within a corporate VPN, ensuring controlled availability and compliance with data privacy requirements.





\subsection{Visualization Layer}

To convert timestamped recognition data into actionable insights, the system employs the Plotly library to generate interactive visualizations. Each visualization is designed not only to present information, but to emphasize patterns in celebrity dynamics and directly support analytical objectives \footnote{https://plotly.com/}.

The layer provides multiple perspectives on the stored data:
\begin{itemize}
\item \textbf{Bar Charts and Pie Charts:} Summarize overall screen time distributions, highlighting the prominence of specific individuals.
\item \textbf{Line Graphs and Stacked Area Charts:} Reveal temporal trends across video timelines, aiding in the study of narrative pacing and character focus.
\item \textbf{Co-Appearance Matrices and Network Graphs:} Illustrate relationships and interactions between celebrities, uncovering clusters and recurring collaborations.
\item \textbf{Heatmaps and Seasonal Comparisons:} Support segment-level intensity analysis and longitudinal comparisons across episodes or seasons, exposing evolving dynamics and long-term trends.
\end{itemize}

\subsection{System Workflow}  

The end-to-end workflow of the system consists of the following stages:  
\begin{enumerate}  
    \item \textbf{Video Processing:} Input videos are processed by the face recognition framework, which detects and identifies individuals in each frame. Timestamped metadata is generated for every detected face.  
    \item \textbf{Data Storage:} The timestamped recognition data is indexed and stored in Elasticsearch, enabling high-speed retrieval and complex queries.  
    \item \textbf{Data Retrieval and Transformation:} Upon user request, relevant data is fetched from Elasticsearch and pre-processed for visualization. This includes aggregation, filtering, and normalization to handle varying video lengths and cast sizes.  
    \item \textbf{Visualization Generation:} The processed data is passed to the visualization layer, which uses Plotly to create interactive visualizations such as bar charts, heatmaps, and network graphs.  
    \item \textbf{Insight Delivery:} The generated visualizations are presented to stakeholders through a dashboard or exported as reports for entertainment analytics, audience studies, and strategic decision-making.  
\end{enumerate}  

This structured workflow ensures seamless integration between data extraction, storage, and visualization, allowing the system to operate efficiently for both small-scale projects and large-scale video datasets. By combining deep learning-based face recognition with powerful visualization techniques, the system bridges the gap between raw video data and actionable insights, offering a comprehensive tool for understanding celebrity dynamics in media content.

\section{Visualization of Celebrity Dynamics}  
  
To evaluate the system's performance and its ability to extract meaningful insights, we applied it to series episodes from Tabii Originals\footnote{https://www.tabii.com/}. The results are presented using ten key visualizations, each highlighting a different aspect of the data. These visualizations are described and analyzed below.  
  
\subsection{Celebrity Appearances per Minute}  
\begin{figure}[!htbp]
    \centering
    \includegraphics[width=0.9\linewidth]{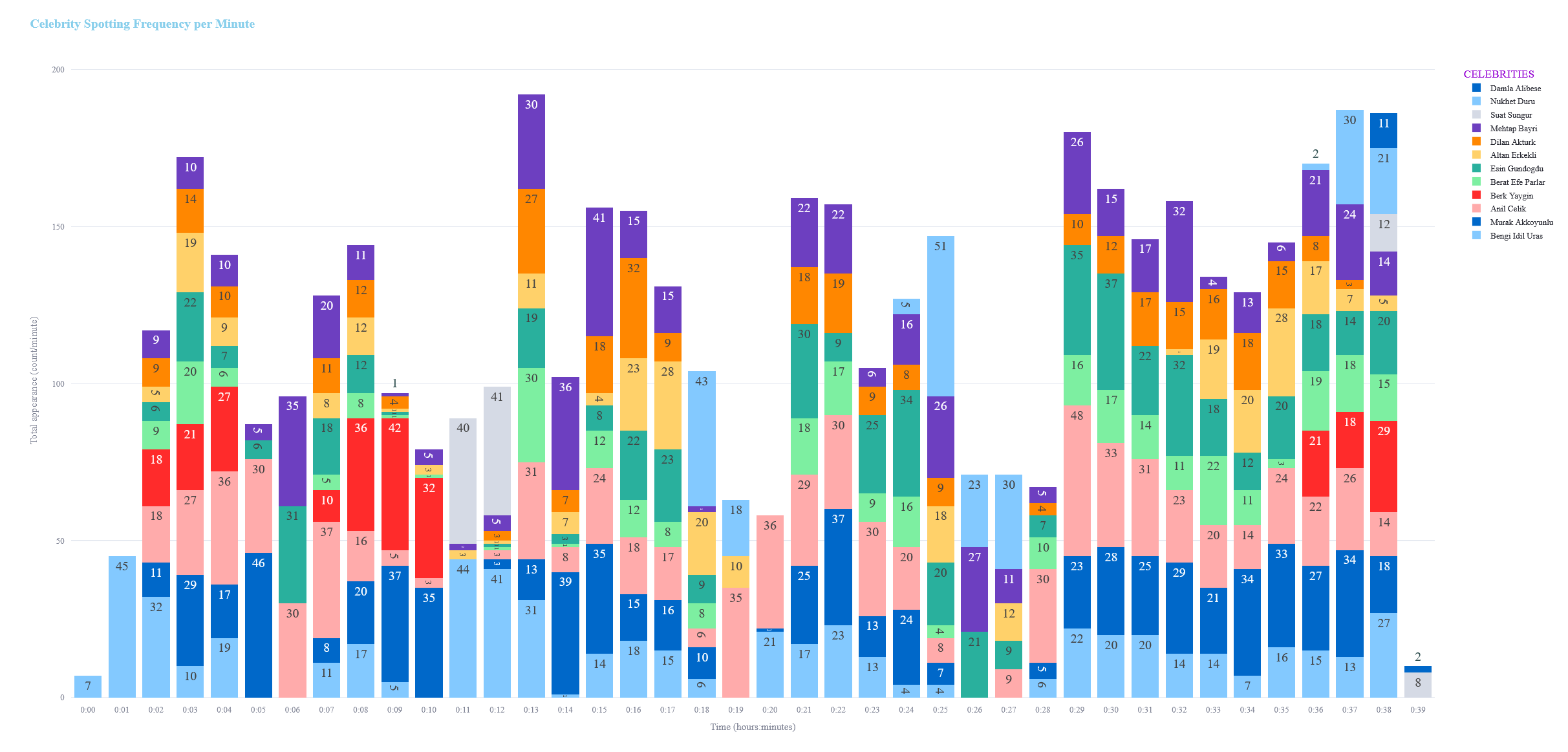}
    \captionsetup{justification=centering, font=small}
    \caption{Celebrity appearances per minute during the episode. 
        The stacked bars represent the number of appearances for each celebrity at a given time. 
        Episode: Ramazan Bey's Mansion, Season 2, Episode 20.}\footnotemark
    \label{fig:appearances_per_minute}
    \vspace{-2mm} 
\end{figure}

Figure~\ref{fig:appearances_per_minute} visualizes the total number of times each celebrity appears within a specific minute of the video. The X-axis represents the time in hours and minutes, while the Y-axis shows the count of appearances for each celebrity during that minute. Each bar in the chart corresponds to a particular minute in the video, and the stacked bars represent the number of appearances for different celebrities.  
  
\footnotetext{IMDb link: \url{https://www.imdb.com/title/tt23108508/}}

\subsection{Total Celebrity Appearance Count}  
\begin{figure}[!htbp]
    \centering
    \includegraphics[width=0.8\linewidth]{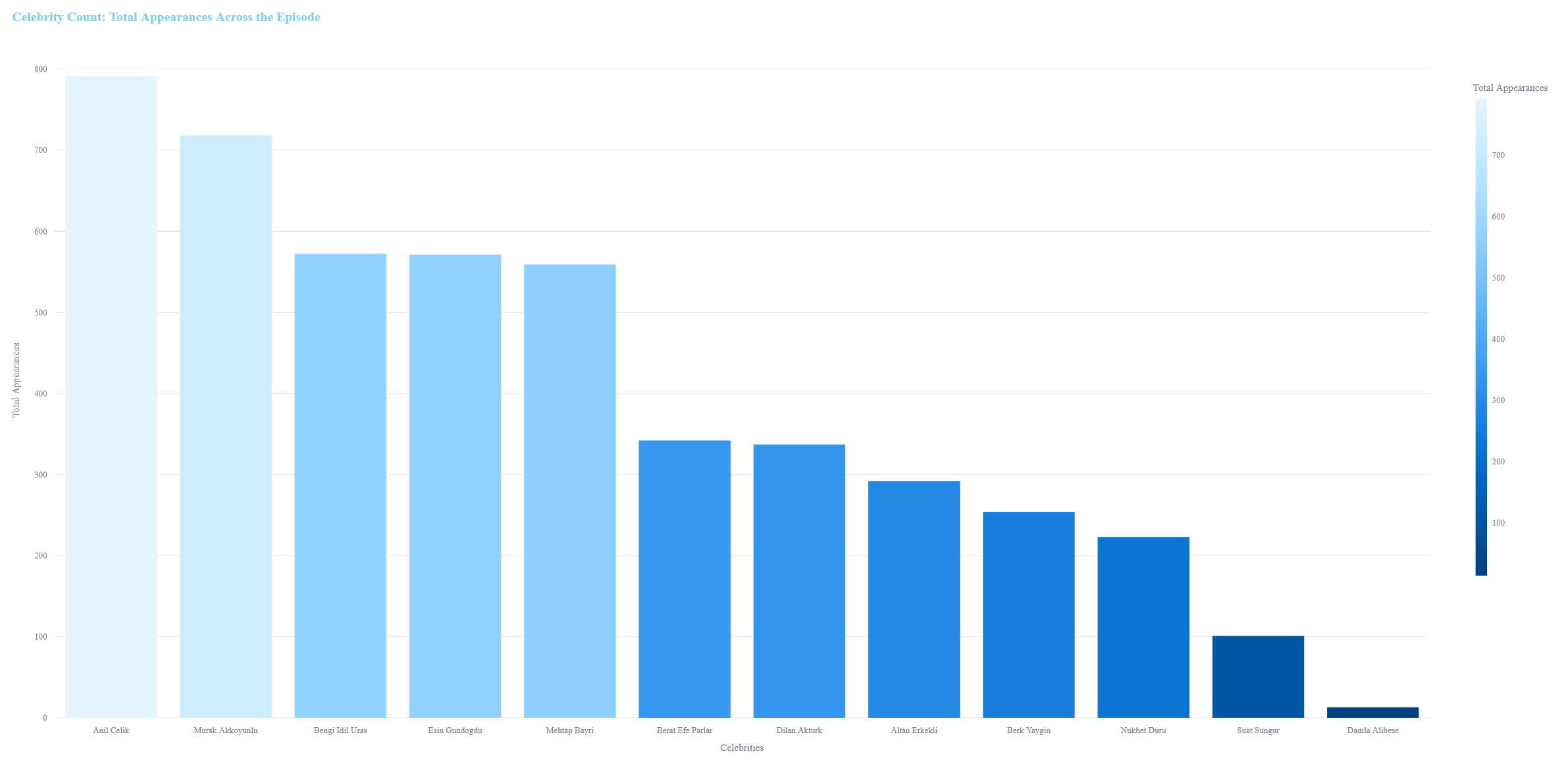}
    \captionsetup{justification=centering, font=small}
    \caption{Total number of appearances for each celebrity throughout the episode. Episode: Ramazan Bey's Mansion, Season 2, Episode 20.}\footnotemark
    \label{fig:total_appearance_count}
    \vspace{-2mm}
\end{figure}

Figure~\ref{fig:total_appearance_count} displays the total number of appearances for each detected celebrity throughout the entire episode. The X-axis lists the names of celebrities, and the Y-axis shows the total count of appearances for each celebrity.  
  
This chart provides a quick overview of which celebrities are the most frequently appearing in the episode. For instance, Celebrity A may dominate the episode with significantly more appearances than others, indicating their central role in the storyline.  
  
\footnotetext{IMDb link: \url{https://www.imdb.com/title/tt23108508/}}
  
\subsection{Total Celebrity Appearance Duration}  
\begin{figure}[!htbp]
    \centering
    \includegraphics[width=0.7\linewidth]{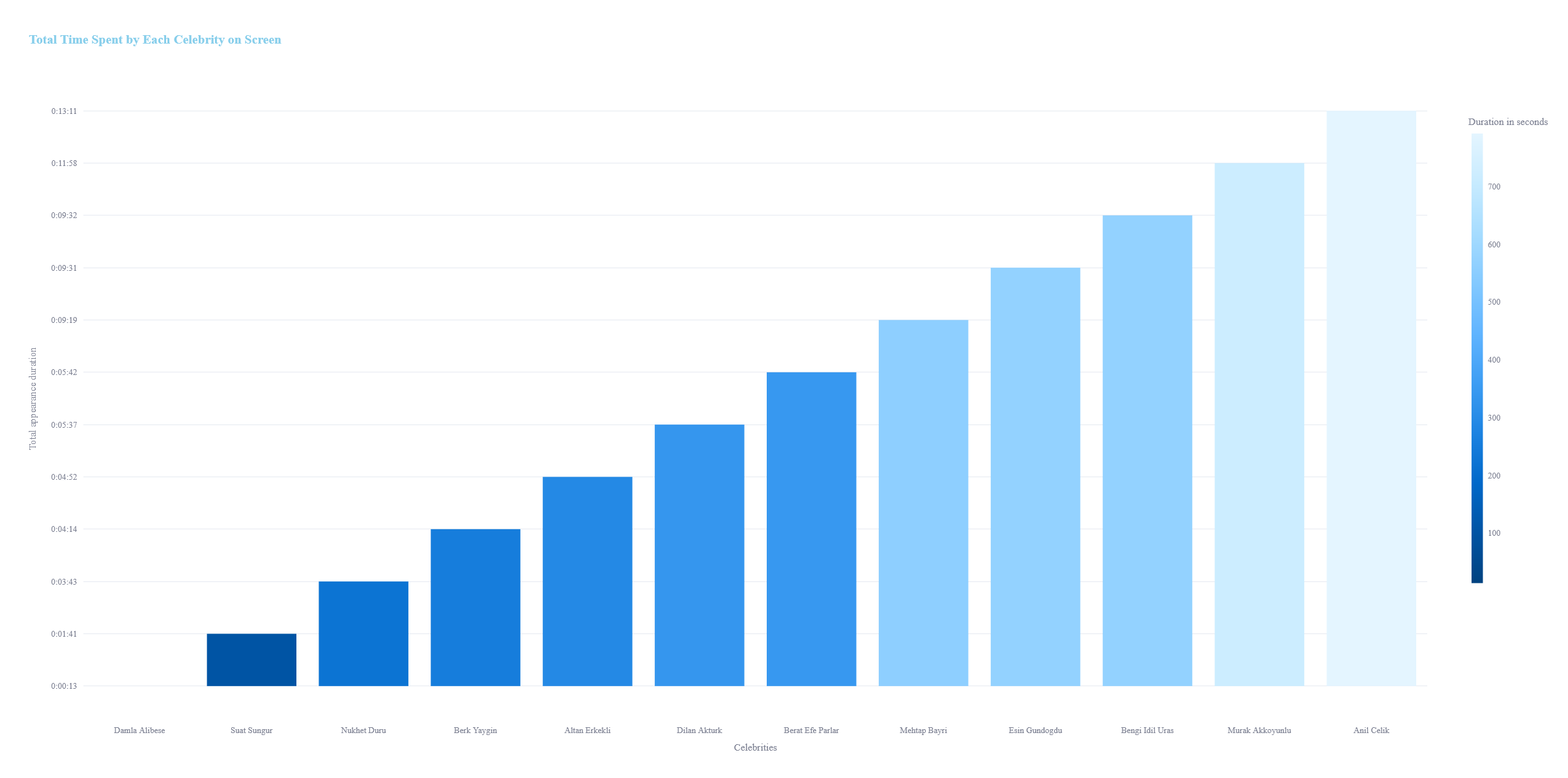}
    \captionsetup{justification=centering, font=small}
    \caption{Total duration of appearances (in hours, minutes, and seconds) for each celebrity during the episode. Episode: Ramazan Bey's Mansion, Season 2, Episode 20.}\footnotemark
    \label{fig:total_appearance_duration}
    \vspace{-2mm}
\end{figure}

Figure~\ref{fig:total_appearance_duration} represents the total duration of each celebrity’s appearance in the video. The Y-axis shows the total time (in hours, minutes, and seconds) each celebrity appeared in the episode. This provides insight into who stayed on screen the longest, emphasizing not just frequency but also prominence.  
  
  
\footnotetext{IMDb link: \url{https://www.imdb.com/title/tt23108508/}}

\subsection{Appearance Trend Over Time}  
\begin{figure}[!htbp]
    \centering
    \includegraphics[width=0.7\linewidth]{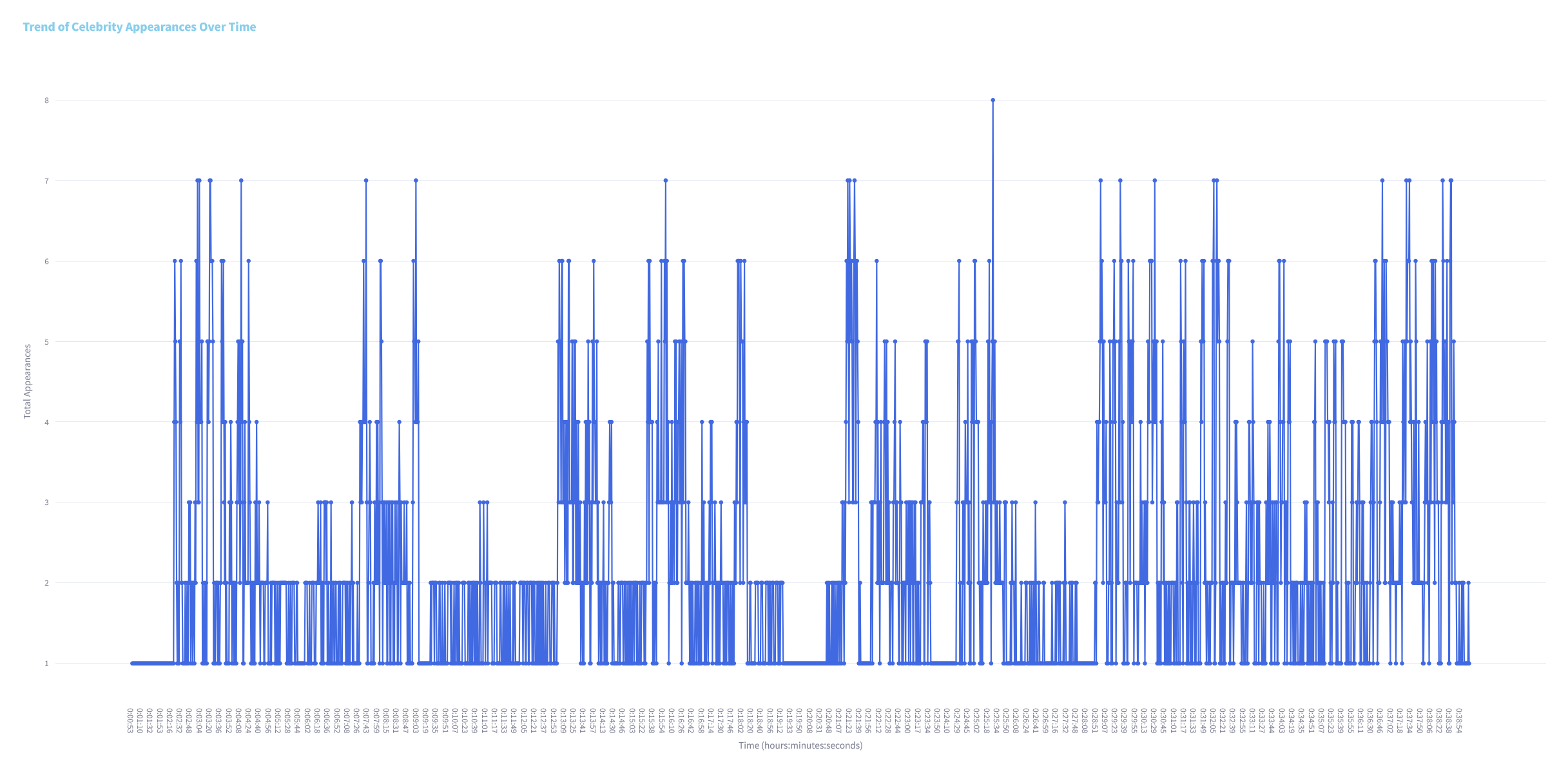}
    \captionsetup{justification=centering, font=small}
    \caption{Appearance trend over time for each celebrity during the episode. Each line represents the count of appearances for a specific celebrity at minute-by-minute intervals. Episode: Ramazan Bey's Mansion, Season 2, Episode 20.}\footnotemark
    \label{fig:appearance_trend}
    \vspace{-2mm} 
\end{figure}

Figure~\ref{fig:appearance_trend} displays the appearance count per minute for each celebrity over the course of the episode. The X-axis represents time in minute-by-minute intervals, while the Y-axis indicates the count of appearances. Each line represents a different celebrity.  
  
  
\footnotetext{IMDb link: \url{https://www.imdb.com/title/tt23108508/}}
  
\subsection{Overall Celebrity Appearance Distribution (Pie Chart)}  
\begin{figure}[!htbp]
    \centering
    \includegraphics[width=0.7\linewidth]{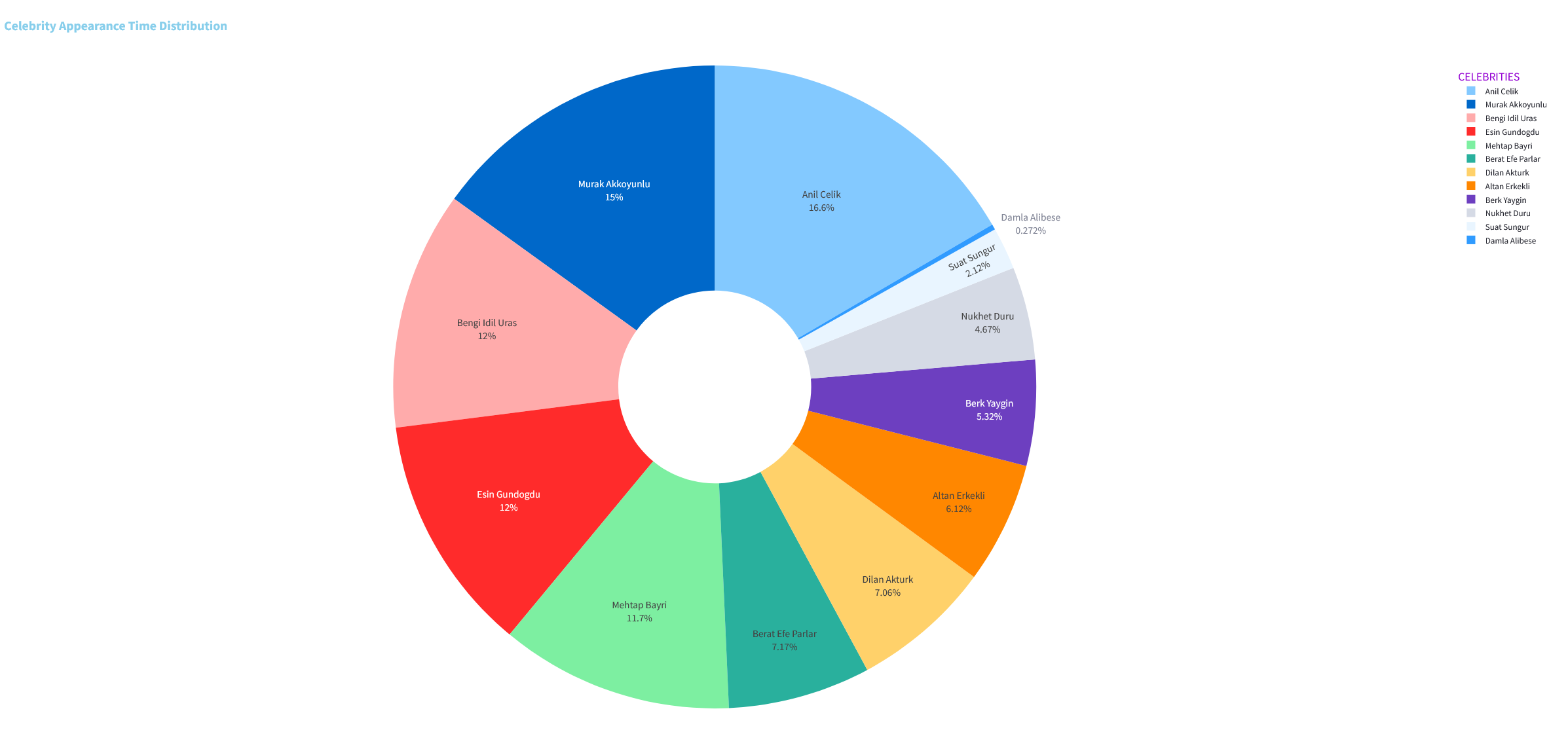}
    \captionsetup{justification=centering, font=small}
    \caption{Pie chart showing the proportional distribution of appearances for each celebrity in the episode. Episode: Ramazan Bey's Mansion, Season 2, Episode 20.}\footnotemark
    \label{fig:appearance_distribution}
    \vspace{-2mm} 
\end{figure}

Figure~\ref{fig:appearance_distribution} shows the proportional distribution of each celebrity’s appearances throughout the episode using a pie chart. Each section represents a celebrity, with the size of the section proportional to their number of appearances.  
  
  
\footnotetext{IMDb link: \url{https://www.imdb.com/title/tt23108508/}}

\subsection{Co-Appearance Matrix}  
\begin{figure}[!htbp]
    \centering
    \includegraphics[width=0.9\linewidth]{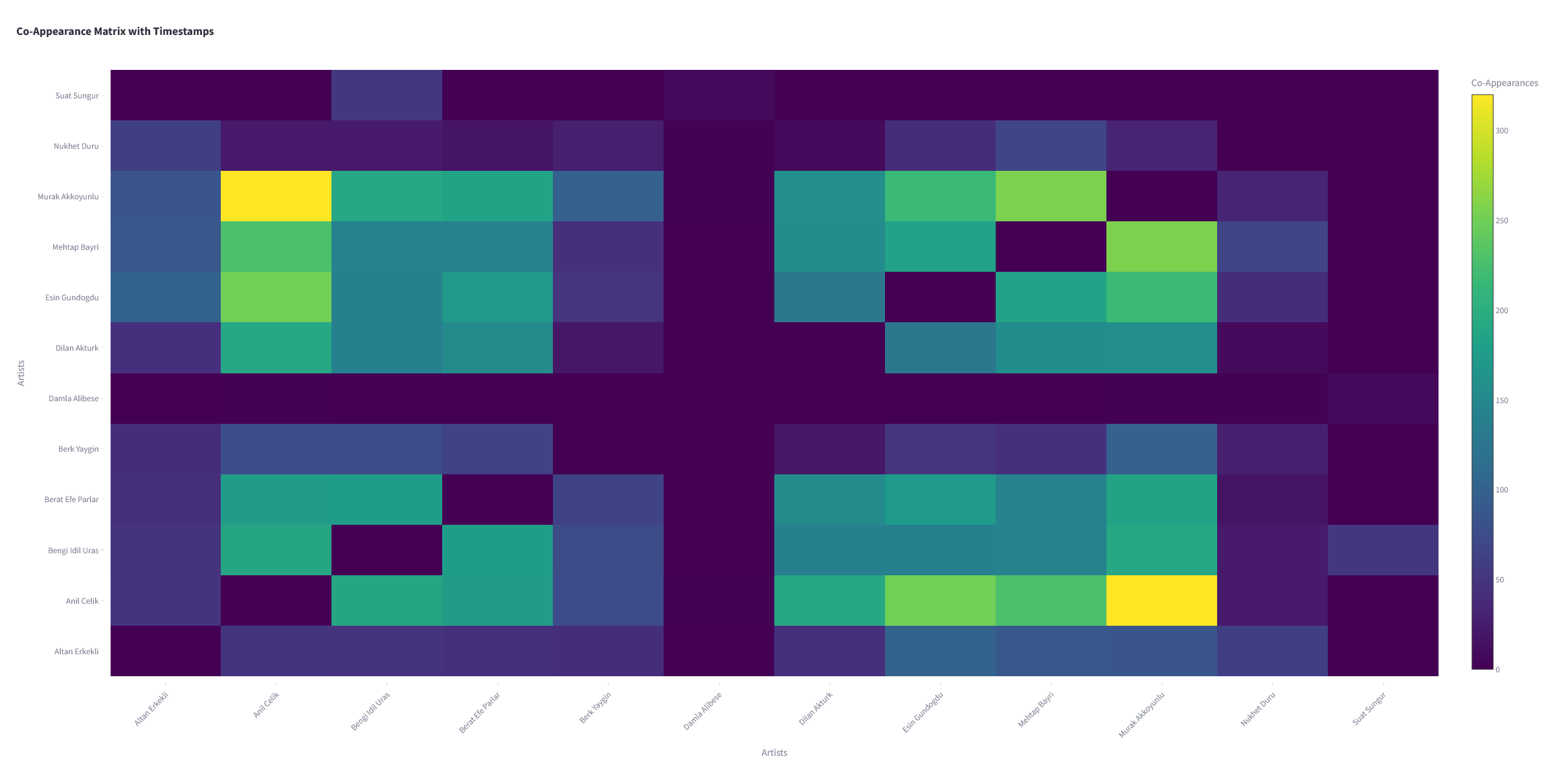}
    \captionsetup{justification=centering, font=small}
    \caption{Co-Appearance Matrix showing the frequency of shared scenes between celebrities during the episode. Episode: Ramazan Bey's Mansion, Season 2, Episode 20.}\footnotemark
    \label{fig:coappearance_matrix}
    \vspace{-2mm} 
\end{figure}

Figure~\ref{fig:coappearance_matrix} visualizes the interactions between celebrities using a heatmap. Each cell represents the frequency of co-appearances between two celebrities during the episode. The X and Y axes list the celebrities, and the color intensity of each cell indicates the number of co-appearances.  
  
  
\footnotetext{IMDb link: \url{https://www.imdb.com/title/tt23108508/}} 
  
\subsection{Co-Appearance Network}  
\begin{figure}[!htbp]
    \centering
    \includegraphics[width=0.9\linewidth]{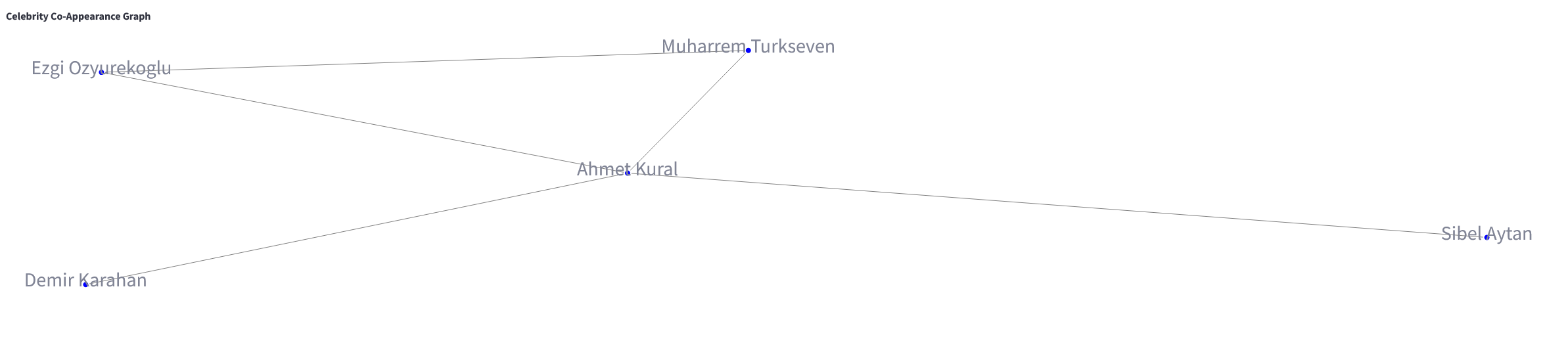}
    \captionsetup{justification=centering, font=small}
    \caption{Co-Appearance Network visualizing the relationships between celebrities during the episode. Episode: The Late Offices, Season 1, Episode 10.}\footnotemark
    \label{fig:coappearance_network}
    \vspace{-2mm} 
\end{figure}

Figure~\ref{fig:coappearance_network} visualizes the relationships between celebrities as a graph. Each node represents a celebrity, and each edge represents a co-appearance between two celebrities. This network visualization highlights the social dynamics within the episode, such as identifying which individuals interact with multiple others. It offers a deeper understanding of the episode’s interaction patterns.  
  
\footnotetext{IMDb link: \url{https://www.imdb.com/title/tt35082593/}} 

\subsection{Stacked Area Chart of Screen Time}
\begin{figure}[!htbp]
    \centering
    \includegraphics[width=0.7\linewidth]{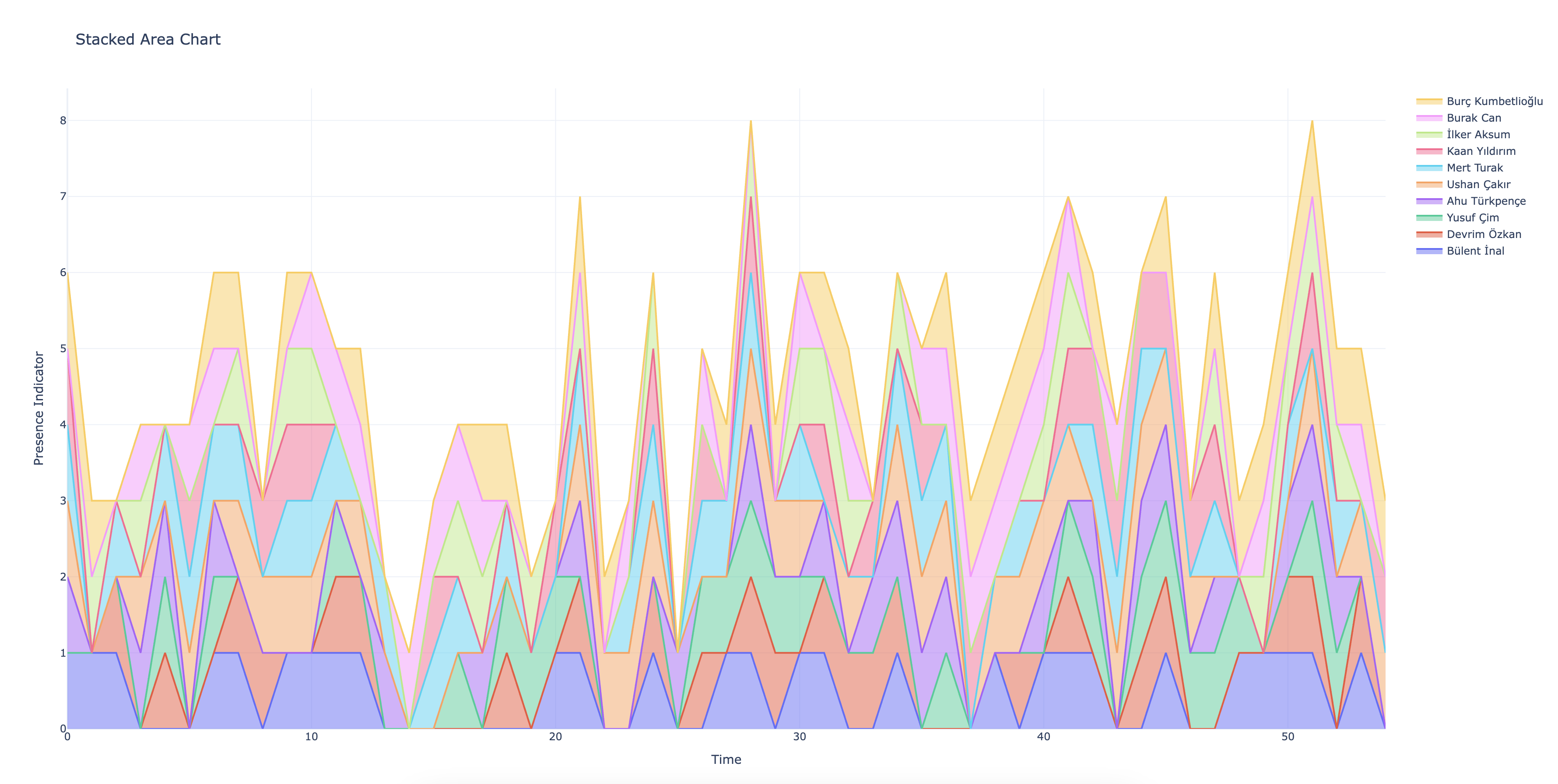}
    \captionsetup{justification=centering, font=small}
    \caption{Cumulative Screen Time Over Episode, Sample from Mevlana Celaleddin-i Rumi, Season 1.}\footnotemark
    \label{fig:stacked_area_chart}
    \vspace{-2mm} 
\end{figure}

Figure~\ref{fig:stacked_area_chart} illustrates the temporal distribution of celebrity appearances throughout the 55-minute episode. The X-axis represents the episode timeline in minutes, while the Y-axis indicates the presence of each celebrity. Each colored layer corresponds to a different celebrity, showing when they appear on screen and how their appearances overlap with others.


\footnotetext{IMDb link: \url{https://www.imdb.com/title/tt14533598/}}

\subsection{Seasonal Comparison Chart}
\begin{figure}[!h]  
  \centering  
  \includegraphics[width=\linewidth]{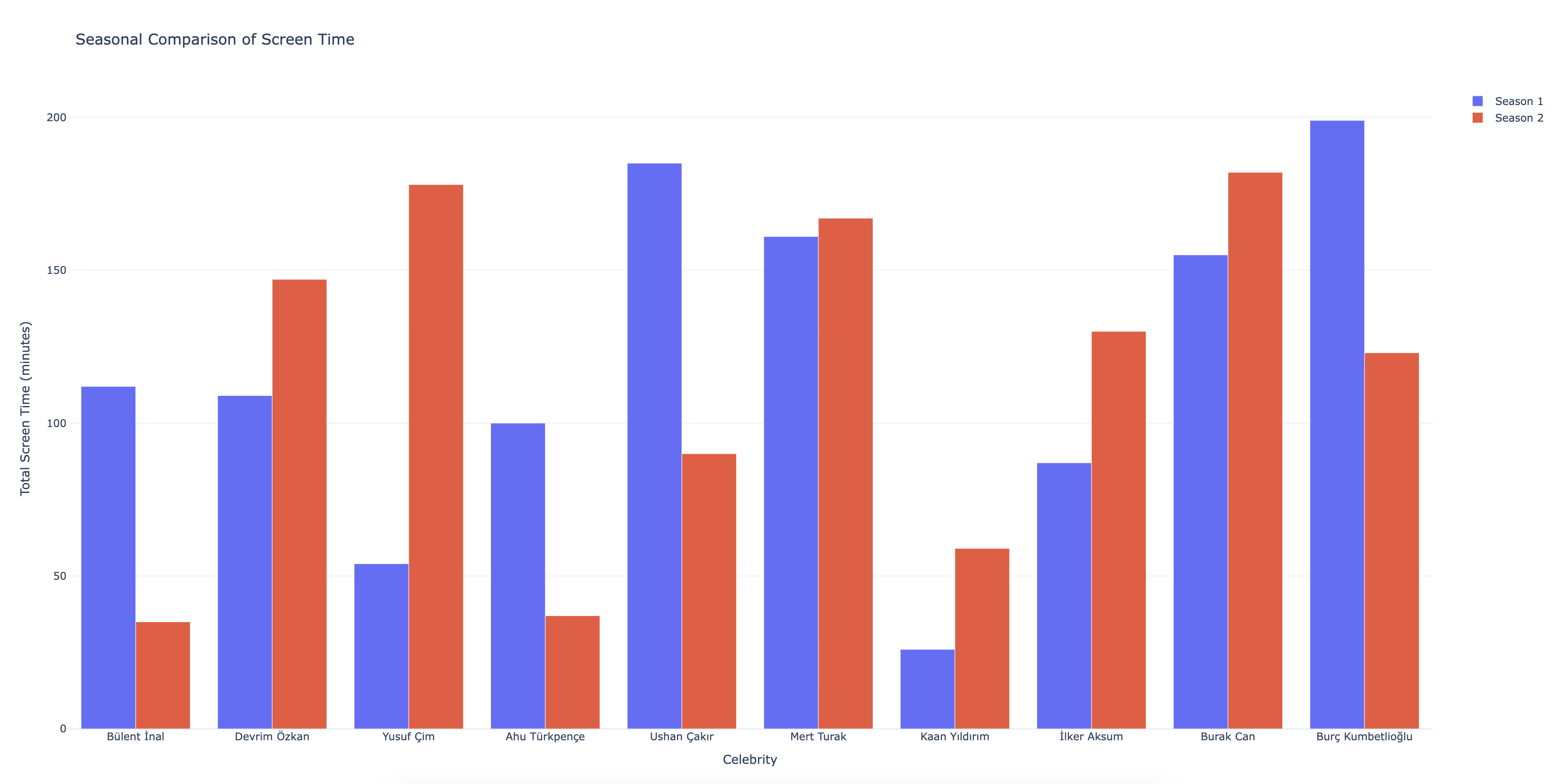}  
  \caption{Screen Time Distribution Across Seasons, Sample from Mevlana Celaleddin-i Rumi, Season 1 and 2.}\footnotemark
  \label{fig:seasonal_comparison}  
\end{figure}

Figure~\ref{fig:seasonal_comparison} presents a comparison of total screen time for each celebrity across two seasons of a show. The X-axis lists the celebrity names, and the Y-axis represents their total screen time in minutes. Bars of different colors correspond to the respective seasons.

This visualization allows for an easy assessment of changes in celebrity prominence between seasons. It can highlight which characters gained or lost focus over time, helping producers and analysts understand long-term storytelling trends and audience engagement shifts. For instance, an increase in screen time for a particular celebrity in Season 2 may indicate their growing narrative importance.

\footnotetext{IMDb link: \url{https://www.imdb.com/title/tt14533598/}}

\subsection{Appearance Heatmap}
\begin{figure}[h]  
  \centering  
  \includegraphics[width=\linewidth]{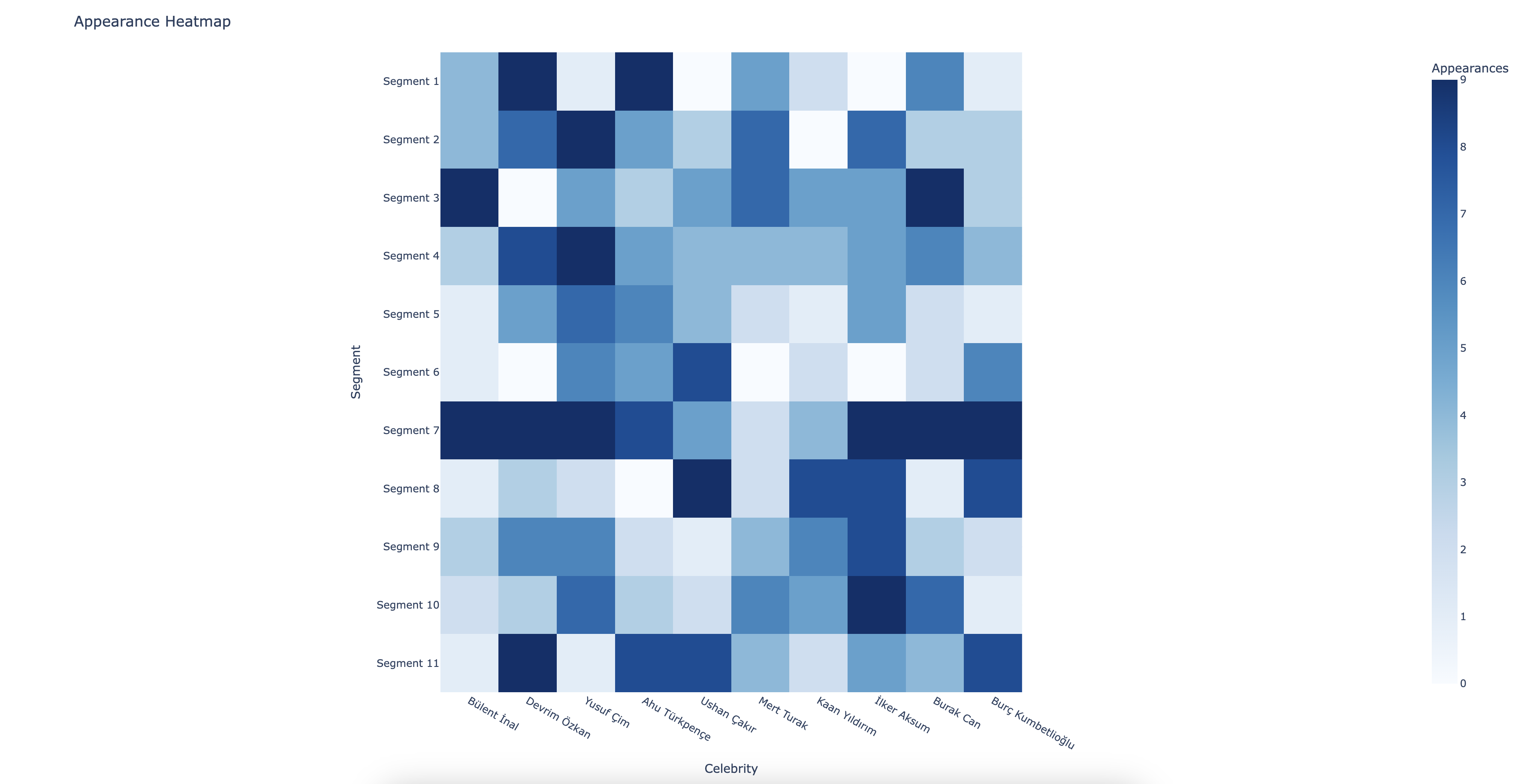}  
  \caption{Appearance Intensity Across Episode Timeline, Sample from Mevlana Celaleddin-i Rumi, Season 1.}\footnotemark  
  \label{fig:heatmap}  
\end{figure}

Figure~\ref{fig:heatmap} shows a segment-wise intensity map of celebrity appearances across the 55-minute episode. The X-axis lists the celebrities, and the Y-axis divides the episode into 11 time segments. Color intensity represents the frequency of appearances within each segment, with darker shades indicating higher presence.

This heatmap allows for a detailed examination of when and how frequently each celebrity appears. It helps identify clusters of high activity, scenes dominated by certain characters, and moments of co-appearance. 

\footnotetext{IMDb link: \url{https://www.imdb.com/title/tt14533598/}}

\newpage

\section{Conclusion}  

In this study, we presented a systematic approach to analyzing celebrity appearances and interactions in video content using a combination of interactive visualizations. By applying this methodology to an example episode, we were able to extract meaningful insights regarding the frequency, duration, and distribution of appearances, as well as the dynamics between different celebrities.  

The visualizations provided complementary perspectives on the data:  
\begin{itemize}  
    \item The \textbf{Celebrity Appearances per Minute} and \textbf{Stacked Area Chart} highlighted temporal trends, moments of high activity, and the evolving focus on different celebrities throughout the episode.  
    \item The \textbf{Total Appearance Count}, \textbf{Duration} charts, and \textbf{Seasonal Comparison} provided a comprehensive overview of each celebrity's prominence and long-term trends across episodes or seasons.  
    \item The \textbf{Co-Appearance Matrix} and \textbf{Network Graphs} revealed relationships and interactions between celebrities, uncovering patterns of collaboration and shared screen time.  
    \item The \textbf{Heatmap} offered segment-wise intensity analysis, enabling detailed examination of when and how frequently each celebrity appears within specific portions of the episode.  
\end{itemize}  

These insights not only help content creators and producers better understand the structure and focus of their media but also provide opportunities to optimize storytelling, pacing, and audience engagement. Furthermore, such analyses can support broader applications, including audience targeting, automated video indexing, and archival organization.  
However, several challenges remain. Detection of celebrities in crowded, poorly lit, or partially occluded scenes can result in inaccuracies, and subtle variations in appearance—such as heavy makeup or accessories—may affect recognition precision. Future work could focus on improving the robustness of face recognition algorithms, incorporating contextual and multi-modal cues. 

In conclusion, this work demonstrates the potential of data-driven video analysis to extract actionable insights about on-screen appearances and relationships. By integrating a richer set of visualizations—including temporal, segment-wise, and seasonal perspectives—the presented framework represents a step forward in automating the understanding of video content, paving the way for deeper insights into media production and consumption.

\newpage
\bibliographystyle{IEEEtran}  
\bibliography{references} 

\begin{thebibliography}{10}
\providecommand{\url}[1]{#1}
\csname url@samestyle\endcsname
\providecommand{\newblock}{\relax}
\providecommand{\bibinfo}[2]{#2}
\providecommand{\BIBentrySTDinterwordspacing}{\spaceskip=0pt\relax}
\providecommand{\BIBentryALTinterwordstretchfactor}{4}
\providecommand{\BIBentryALTinterwordspacing}{\spaceskip=\fontdimen2\font plus
\BIBentryALTinterwordstretchfactor\fontdimen3\font minus \fontdimen4\font\relax}
\providecommand{\BIBforeignlanguage}[2]{{%
\expandafter\ifx\csname l@#1\endcsname\relax
\typeout{** WARNING: IEEEtran.bst: No hyphenation pattern has been}%
\typeout{** loaded for the language `#1'. Using the pattern for}%
\typeout{** the default language instead.}%
\else
\language=\csname l@#1\endcsname
\fi
#2}}
\providecommand{\BIBdecl}{\relax}
\BIBdecl

\bibitem{steed2025interactive}
C.~A. Steed, ``Interactive data visualization,'' in \emph{Data Analytics for Intelligent Transportation Systems}.\hskip 1em plus 0.5em minus 0.4em\relax Elsevier, 2025, pp. 185--211.

\bibitem{bitter2007comparison}
I.~Bitter, R.~Van~Uitert, I.~Wolf, L.~Ib{\'a}{\~n}ez, and J.-M. Kuhnigk, ``Comparison of four freely available frameworks for image processing and visualization that use itk,'' \emph{IEEE transactions on visualization and computer graphics}, vol.~13, no.~3, pp. 483--493, 2007.

\bibitem{kindlmann2015diderot}
G.~Kindlmann, C.~Chiw, N.~Seltzer, L.~Samuels, and J.~Reppy, ``Diderot: a domain-specific language for portable parallel scientific visualization and image analysis,'' \emph{IEEE transactions on visualization and computer graphics}, vol.~22, no.~1, pp. 867--876, 2015.

\bibitem{lai2017semantic}
W.-S. Lai, Y.~Huang, N.~Joshi, C.~Buehler, M.-H. Yang, and S.~B. Kang, ``Semantic-driven generation of hyperlapse from 360 degree video,'' \emph{IEEE transactions on visualization and computer graphics}, vol.~24, no.~9, pp. 2610--2621, 2017.

\bibitem{wang2013spatially}
Y.-S. Wang, F.~Liu, P.-S. Hsu, and T.-Y. Lee, ``Spatially and temporally optimized video stabilization,'' \emph{IEEE transactions on visualization and computer graphics}, vol.~19, no.~8, pp. 1354--1361, 2013.

\bibitem{xiao2010fast}
C.~Xiao, M.~Liu, N.~Yongwei, and Z.~Dong, ``Fast exact nearest patch matching for patch-based image editing and processing,'' \emph{IEEE Transactions on Visualization and Computer Graphics}, vol.~17, no.~8, pp. 1122--1134, 2010.

\bibitem{zeng2019emoco}
H.~Zeng, X.~Wang, A.~Wu, Y.~Wang, Q.~Li, A.~Endert, and H.~Qu, ``Emoco: Visual analysis of emotion coherence in presentation videos,'' \emph{IEEE transactions on visualization and computer graphics}, vol.~26, no.~1, pp. 927--937, 2019.

\bibitem{botchen2008action}
R.~P. Botchen, S.~Bachthaler, F.~Schick, M.~Chen, G.~Mori, D.~Weiskopf, and T.~Ertl, ``Action-based multifield video visualization,'' \emph{IEEE Transactions on Visualization and Computer Graphics}, vol.~14, no.~4, pp. 885--899, 2008.

\bibitem{chan2019motion}
G.~Y.-Y. Chan, L.~G. Nonato, A.~Chu, P.~Raghavan, V.~Aluru, and C.~T. Silva, ``Motion browser: visualizing and understanding complex upper limb movement under obstetrical brachial plexus injuries,'' \emph{IEEE Transactions on Visualization and Computer Graphics}, vol.~26, no.~1, pp. 981--990, 2019.

\bibitem{fan2019vista}
M.~Fan, K.~Wu, J.~Zhao, Y.~Li, W.~Wei, and K.~N. Truong, ``Vista: Integrating machine intelligence with visualization to support the investigation of think-aloud sessions,'' \emph{IEEE Transactions on Visualization and Computer Graphics}, vol.~26, no.~1, pp. 343--352, 2019.

\bibitem{jang2015motionflow}
S.~Jang, N.~Elmqvist, and K.~Ramani, ``Motionflow: Visual abstraction and aggregation of sequential patterns in human motion tracking data,'' \emph{IEEE transactions on visualization and computer graphics}, vol.~22, no.~1, pp. 21--30, 2015.

\bibitem{lu2012timeline}
S.-P. Lu, S.-H. Zhang, J.~Wei, S.-M. Hu, and R.~R. Martin, ``Timeline editing of objects in video,'' \emph{IEEE Transactions on Visualization and Computer Graphics}, vol.~19, no.~7, pp. 1218--1227, 2012.

\bibitem{meghdadi2013interactive}
A.~H. Meghdadi and P.~Irani, ``Interactive exploration of surveillance video through action shot summarization and trajectory visualization,'' \emph{IEEE Transactions on Visualization and Computer Graphics}, vol.~19, no.~12, pp. 2119--2128, 2013.

\bibitem{romero2008viz}
M.~Romero, J.~Summet, J.~Stasko, and G.~Abowd, ``Viz-a-vis: Toward visualizing video through computer vision,'' \emph{IEEE Transactions on Visualization and Computer Graphics}, vol.~14, no.~6, pp. 1261--1268, 2008.

\bibitem{wang2017outdoor}
Y.~Wang, Y.~Liu, X.~Tong, Q.~Dai, and P.~Tan, ``Outdoor markerless motion capture with sparse handheld video cameras,'' \emph{IEEE transactions on visualization and computer graphics}, vol.~24, no.~5, pp. 1856--1866, 2017.

\bibitem{celebvhq2022}
H.~Zhu, W.~Wu, W.~Zhu, L.~Jiang, S.~Tang, L.~Zhang, Z.~Liu, and C.~C. Loy, ``Celebv-hq: A large-scale video facial attributes dataset,'' \emph{arXiv preprint arXiv:2207.12393}, 2022.

\bibitem{facevid1k2024}
A.~Authors], ``Facevid-1k: A large-scale high-quality multiracial human face video dataset,'' \emph{arXiv preprint arXiv:2410.07151}, 2024.

\bibitem{duet2025}
L.~Xie, K.~Xu, J.~Zhao, and M.~Chen, ``Duet: Exploring event visualizations on timelines,'' in \emph{Proceedings of the ACM CHI Conference on Human Factors in Computing Systems}.\hskip 1em plus 0.5em minus 0.4em\relax ACM, 2025.

\bibitem{adaptive360timeline2024}
Z.~Liu, J.~Li, and A.~Smith, ``Adaptive timeline designs for 360° video exploration in virtual reality,'' \emph{Computers \& Graphics}, vol. 116, 2024.

\bibitem{vistellar2024}
W.~Zhang, Y.~Cao, and X.~Wang, ``Vistellar: Embedding visualizations into short-form videos with post-recording timelines,'' in \emph{Proceedings of IEEE VIS}, 2024.

\bibitem{dataplaywright2024}
H.~Kim, Y.~Chen, and J.~Heer, ``Data playwright: Authoring data videos with synchronized narration and animation,'' \emph{arXiv preprint arXiv:2406.12345}, 2024.

\bibitem{chen2015peakvizor}
Q.~Chen, Y.~Chen, D.~Liu, C.~Shi, Y.~Wu, and H.~Qu, ``Peakvizor: Visual analytics of peaks in video clickstreams from massive open online courses,'' \emph{IEEE transactions on visualization and computer graphics}, vol.~22, no.~10, pp. 2315--2330, 2015.

\bibitem{polk2014tennivis}
T.~Polk, J.~Yang, Y.~Hu, and Y.~Zhao, ``Tennivis: Visualization for tennis match analysis,'' \emph{IEEE transactions on visualization and computer graphics}, vol.~20, no.~12, pp. 2339--2348, 2014.

\bibitem{polk2019courttime}
T.~Polk, D.~J{\"a}ckle, J.~H{\"a}u{\ss}ler, and J.~Yang, ``Courttime: Generating actionable insights into tennis matches using visual analytics,'' \emph{IEEE Transactions on Visualization and Computer Graphics}, vol.~26, no.~1, pp. 397--406, 2019.

\bibitem{wang2018image}
Y.~Wang, Z.~Wang, C.-W. Fu, H.~Schmauder, O.~Deussen, and D.~Weiskopf, ``Image-based aspect ratio selection,'' \emph{IEEE Transactions on Visualization and Computer Graphics}, vol.~25, no.~1, pp. 840--849, 2018.

\bibitem{liao2011video}
M.~Liao, J.~Gao, R.~Yang, and M.~Gong, ``Video stereolization: Combining motion analysis with user interaction,'' \emph{IEEE Transactions on Visualization and Computer Graphics}, vol.~18, no.~7, pp. 1079--1088, 2011.

\bibitem{liu2009point}
Y.~Liu, Q.~Dai, and W.~Xu, ``A point-cloud-based multiview stereo algorithm for free-viewpoint video,'' \emph{IEEE transactions on visualization and computer graphics}, vol.~16, no.~3, pp. 407--418, 2009.

\bibitem{serrano2019motion}
A.~Serrano, I.~Kim, Z.~Chen, S.~DiVerdi, D.~Gutierrez, A.~Hertzmann, and B.~Masia, ``Motion parallax for 360 rgbd video,'' \emph{IEEE Transactions on Visualization and Computer Graphics}, vol.~25, no.~5, pp. 1817--1827, 2019.

\bibitem{penta2025}
H.~Sun, J.~Yang, D.~Holten, and T.~Dwyer, ``Penta: Visualizing compound graphs as set-typed data,'' in \emph{Proceedings of IEEE PacificVis}, 2025.

\bibitem{filmsurvey2024}
T.~Peixoto and A.~Garas, ``Leading by the nodes: A survey of film industry network analysis,'' \emph{Applied Network Science}, vol.~9, no.~1, p.~45, 2024.

\bibitem{actorcollab2024}
R.~Gupta and A.~Kumar, ``Analyzing social networks of actors in movies and tv series,'' \emph{arXiv preprint arXiv:2405.09876}, 2024.

\bibitem{unicooccurrence2025}
X.~Li, Q.~Wang, and M.~Zhao, ``Building and visualizing digital-media co-occurrence networks of universities,'' \emph{Humanities and Social Sciences Communications}, vol.~12, no.~77, 2025.

\bibitem{pnascortex2024}
Y.~Chen, S.~Nastase, and U.~Hasson, ``Cortico-hippocampal networks carry information about characters during naturalistic viewing,'' \emph{PNAS Nexus}, vol.~3, no.~5, p. pgad123, 2024.

\bibitem{arcvideo2025}
L.~Wang, C.~Zhou, and R.~Zhang, ``Arc-hunyuan-video-7b: Structured video comprehension with temporal precision,'' \emph{arXiv preprint arXiv:2501.04567}, 2025.

\bibitem{coact2024}
A.~Rahman, P.~Singh, and H.~Chen, ``Co-act: A dataset for human action co-occurrence in videos,'' \emph{arXiv preprint arXiv:2403.07654}, 2024.

\bibitem{cagnet2024}
F.~Zhou, L.~Huang, and R.~Xu, ``Cagnet: Video social relationship graph generation via co-occurrence aware graph networks,'' in \emph{Proceedings of the IEEE/CVF Conference on Computer Vision and Pattern Recognition}, 2024.

\bibitem{ehcgcn2025}
X.~Tang, Z.~Li, and K.~Wu, ``Ehc-gcn: Efficient hierarchical co-occurrence graph convolutional network for action recognition,'' \emph{IEEE Transactions on Multimedia}, 2025.

\bibitem{10581926}
D.~Demir and I.~D. Elkahlout, ``A study on end-to-end face analysis: How to cope with challenges,'' in \emph{2024 IEEE 18th International Conference on Automatic Face and Gesture Recognition (FG)}, 2024, pp. 1--10.

\end{thebibliography}
\end{document}